\documentclass[letterpaper, 10 pt, conference]{ieeeconf}  

\usepackage{cite}
\usepackage{tikz}
\usepackage{graphicx}
\usetikzlibrary{shapes.geometric, arrows}
\usepackage{amsmath}

\usepackage{multirow}

\IEEEoverridecommandlockouts                              

\title{\LARGE \bf
3D Fully Convolutional Network for Vehicle Detection in Point Cloud
}

\author{Bo Li*
\thanks{*Bo Li is a researcher at Baidu Inc. Contact: {\tt prclibo.github.io} or {\tt libo24@baidu.com}}%
}

\begin{document}

\maketitle
\thispagestyle{empty}
\pagestyle{empty}

\begin{abstract}
2D fully convolutional network has been recently successfully applied to object detection from images. In this paper, we extend the fully convolutional network based detection techniques to 3D and apply it to point cloud data. The proposed approach is verified on the task of vehicle detection from lidar point cloud for autonomous driving. Experiments on the KITTI dataset shows a significant performance improvement over the previous point cloud based detection approaches. 

\end{abstract}

\section{INTRODUCTION}

Understanding point cloud data has been recognized as an inevitable task for many robotic applications. Compared to image based detection, object detection in point cloud naturally localizes the 3D coordinates of the objects, which provides crucial information for subsequent tasks like navigation or manipulation.

In this paper, we design a 3D fully convolutional network (FCN) to detect and localize objects as 3D boxes from point cloud data. The 2D FCN \cite{Long} has achieved notable performance in image based detection tasks. The proposed approach extends FCN to 3D and is applied to 3D vehicle detection for an autonomous driving system, using a Velodyne 64E lidar. Meanwhile, the approach can be generalized to other object detection tasks on point cloud captured by Kinect, stereo or monocular structure from motion.

\section{RELATED WORKS}

\subsection{3D Object Detection in Point Cloud}
A majority of 3D detection algorithms can be summarized as two stages, i.e. candidate proposal and classification. Candidates can be proposed by delicate segmentation algorithms \cite{Himmelsbach2010, Moosmann2009, Douillard2011, Wang2012, Klasing2008, Papon2013, Triebel2006, Triebel2005, Behley2013a}, sliding window \cite{Wang}, random sampling \cite{Johnson1999}, or the recently popular Region Proposal Network (RPN) \cite{Song2014}. For the classification stage, research have been drawn to features including shape model \cite{Faugeras1986, Johnson1999} and geometry statistic features \cite{Triebel2006, Wang2012, Teichman2011, Papon2013}. Sparsing coding \cite{Deuge2013, Lai2014} and deep learning \cite{Song2014} are also used for feature representation.

Besides directly operating in the 3D point cloud space, some other previous detection alogrithms project 3D point cloud onto 2D surface as depthmaps or range scans \cite{Chen2015, Lin2013, Li2016}. The projection inevitably loses or distorts useful 3D spatial information but can benefit from the well developed image based 2D detection algorithms.

\subsection{Convolutional Neural Network and 3D Object Detection}

CNN based 3D object detection is recently drawing a growing attention in computer vision and robotics. \cite{Chen2015, Lin2013, Li2016, Gupta2014, Schwarz2015, Socher2012} embed 3D information in 2D projection and use 2D CNN for recognition or detection. \cite{Li2016} also suggest it possible to predict 3D object localization by 2D CNN network on range scans. \cite{HegdeStanford} operates 3D voxel data but regards one dimension as a channel to apply 2D CNN. \cite{Wu2015, Maturana2015, Graham2015, Song2014} are among the very few earlier works on 3D CNN. \cite{Wu2015, Maturana2015, Graham2015} focus on object recognition and \cite{Song2014} proposes 3D R-CNN techniques for indoor object detection combining the Kinect image and point cloud.

\hfill \break

In this paper, we transplant the fully convolutional network (FCN) to 3D to detect and localize object as 3D boxes in point cloud. The FCN is a recently popular framework for end-to-end object detection, with top performance in tasks including ImageNet, KITTI, ICDAR, etc. Variations of FCN include DenseBox \cite{Huang2015}, YOLO \cite{Redmon2015} and SSD \cite{Liu2015}. The approach proposed in this paper is inspired by the basic idea of DenseBox.

\begin{figure}
    \centering
    \includegraphics[scale=0.8]{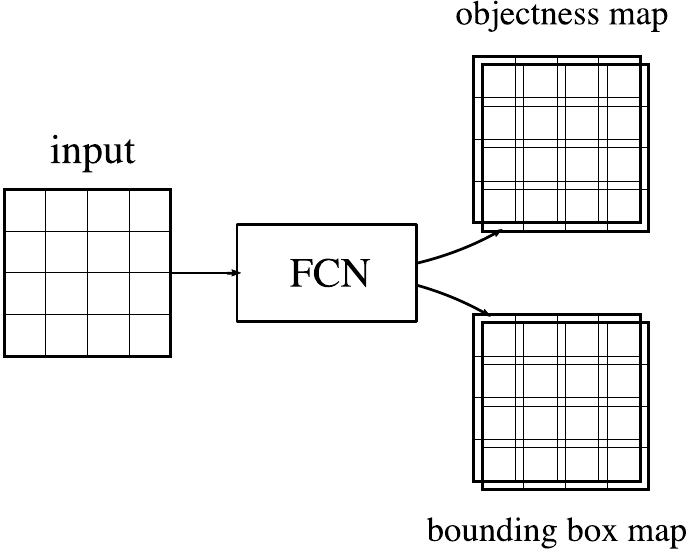}
    \caption{A sample illustration of the structure of the FCN.}
    \label{fig:fcn2d-net}
\end{figure}

\begin{figure*}
    \centering
    \includegraphics[scale=0.8]{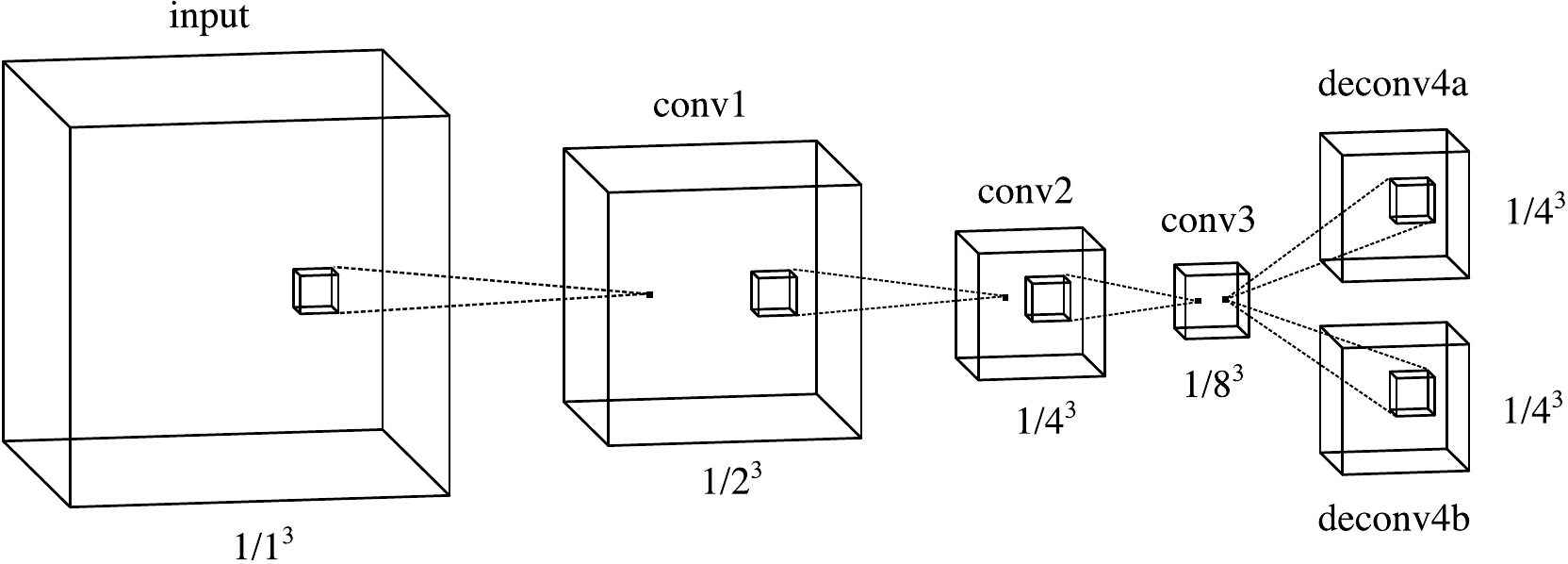}
    \caption{A sample illustration of the 3D FCN structure used in this paper. Feature maps are first down-sampled by three convolution operation with the stride of $1 / 2^3$ and then up-samped by the deconvolution operation of the same stride. The output objectness map ($\mathbf{o}^a$) and bounding box map ($\mathbf{o}^b$) are collected from the deconv4a and deconv4b layers respectively.
}
    \label{fig:fcn3d-net}
\end{figure*}

\section{APPROACH}

\subsection{FCN Based Detection Revisited}
The procedure of FCN based detection frameworks can be summarized as two tasks, i.e. objectness prediction and bounding box prediction. As illustrated in Figure \ref{fig:fcn2d-net}, a FCN is formed with two output maps corresponding to the two tasks respectively. The objectness map predicts if a region belongs to an object and the bounding box map predicts the coordinates of the object bounding box. We follow the denotion of \cite{Li2016}. Denote $\mathbf{o}^a_\mathbf{p}$ as the output at region $\mathbf{p}$ of the objectness map, which can be encoded by softmax or hinge loss. Denote $\mathbf{o}^b_\mathbf{p}$ as the output of the bounding box map, which is encoded by the coordinate offsets of the bounding box. 

Denote the groundtruth objectness label at region $\mathbf{p}$ as $\ell_\mathbf{p}$. For simplicity each class corresponds to one label in this paper. In some works, e.g. SSD or DenseBox, the network can have multiple objectness labels for one class, corresponding to multiple scales or aspect ratios. The objectness loss at $\mathbf{p}$ is denoted as
\begin{equation}
\begin{aligned}
    \mathcal{L}_\textrm{obj}(\mathbf{p}) &= - \log (p_\mathbf{p}) \\
    p_\mathbf{p} &= \cfrac{\exp (-\mathbf{o}^a_{\mathbf{p}, \ell_\mathbf{p}})}{ \sum_{\ell \in \{0, 1\}}{\exp (-\mathbf{o}^a_{\mathbf{p}, \ell})}}
\end{aligned}
\label{eq:object_loss}
\end{equation}

Denote the groundtruth bounding box coordinates offsets at region $\mathbf{p}$ as $\mathbf{b}_\mathbf{p}$. Similarly, in this paper we assume only one bounding box map is produced, though a more sophisticated network can have multiple bounding box offsets predicted for one class, corresponding to multiple scales or aspect ratios. Each bounding box loss is denoted as
\begin{equation}
\begin{aligned}
    \mathcal{L}_\textrm{box}(\mathbf{p}) = \| \mathbf{o}^b_{\mathbf{p}} - \mathbf{b}_\mathbf{p} \|^2
\end{aligned}
\label{eq:box_loss}
\end{equation}

The overall loss of the network is thus denoted as
\begin{equation}
    \mathcal{L} = \sum_{\mathbf{p} \in \mathcal{P}} \mathcal{L}_\textrm{obj}(\mathbf{p}) + w \sum_{\mathbf{p} \in \mathcal{V}} \mathcal{L}_\textrm{box}(\mathbf{p})
\end{equation}
with $w$ used to balance the objectness loss and the bounding box loss. $\mathcal{P}$ denotes all regions in the objectness map and $\mathcal{V} \in \mathcal{P}$ denotes all object regions. In the deployment phase, the regions with postive objectness prediction are selected. Then the bounding box predictions corresponding to these regions are collected and clustered as the detection results.

\begin{figure*}
    \centering
    \begin{tabular}{ccc}
        \includegraphics[scale=0.35, trim=70 50 50 40, clip]{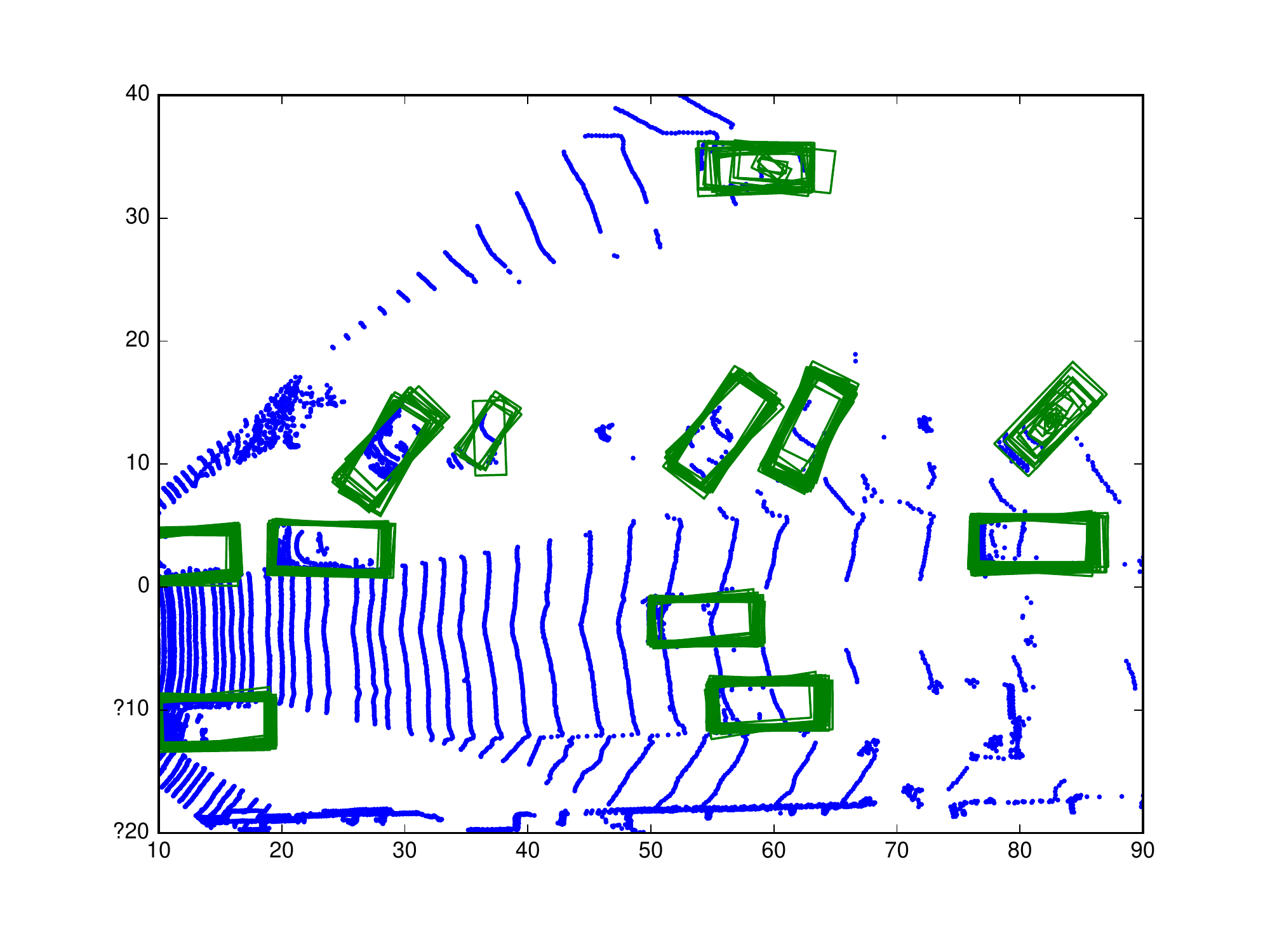} & \includegraphics[scale=0.35, trim=70 50 50 40, clip]{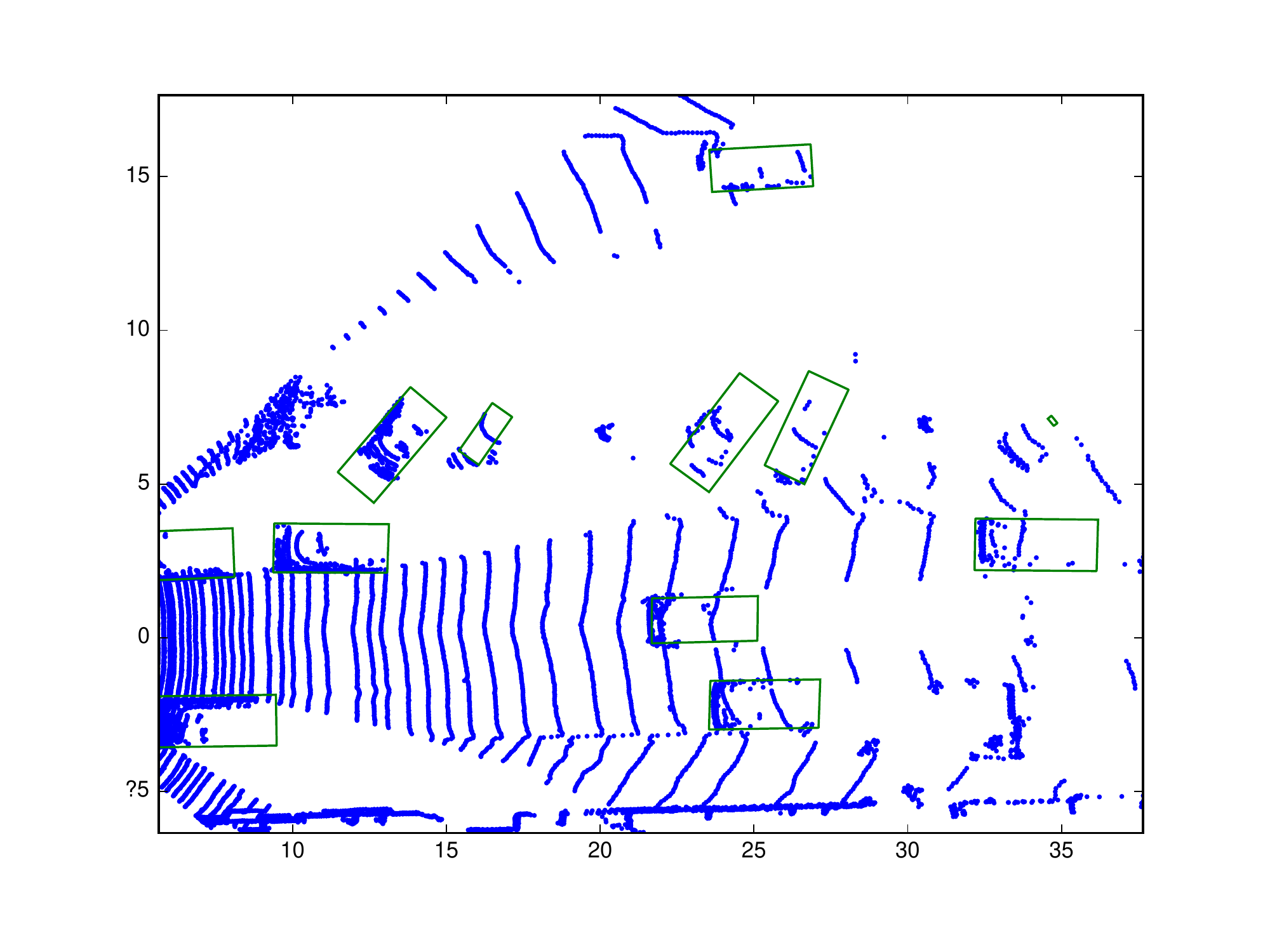} &
        \includegraphics[scale=0.35, trim=50 50 30 40, clip]{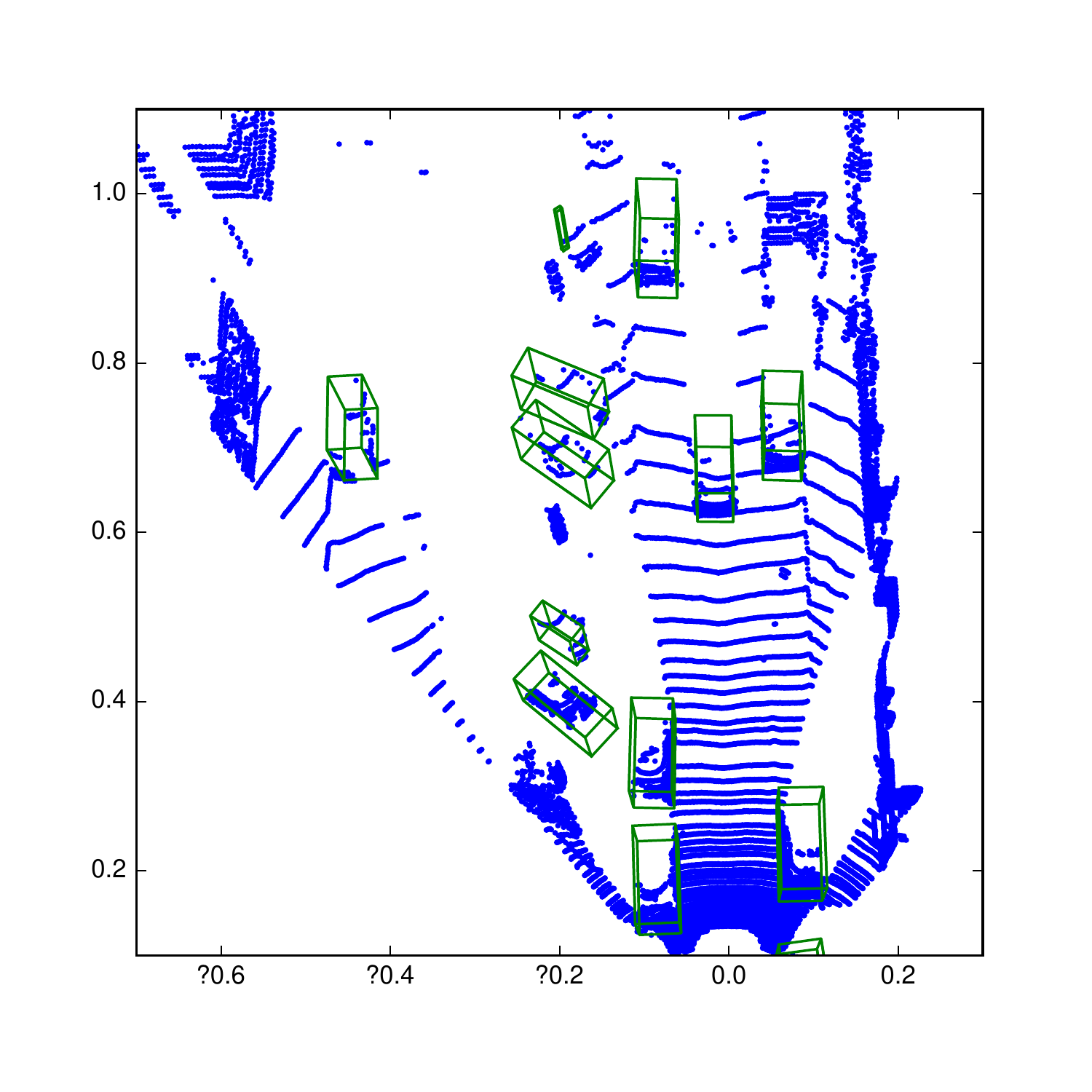}\\
        (a) & (b) & (c)
    \end{tabular}
    \caption{Intermediate results of the 3D FCN detection procedure. (a) Bounding box predictions are collected from regions with high objectness confidence and are plotted as green boxes. (b) Bounding boxes after clustering plotted with the blue original point cloud. (c) Detection in 3D since (a) and (b) are visualized in the bird's eye view. }
    \label{fig:medium-output}
\end{figure*}

\subsection{3D FCN Detection Network for Point Cloud}

Although a variety of discretization embedding have been introduced for high-dimensional convolution \cite{Adams2010, Graham2015}, for simplicity we discretize the point cloud on square grids. The discretized data can be represented by a 4D array with dimensions of length, width, height and channels. For the simplest case, only one channel of value $\{0, 1\}$ is used to present whether there is any points observed at the corresponding grid elements. Some more sophisticated features have also been introduced in the previous works, e.g. \cite{Maturana2015}.

The mechanism of 2D CNN naturally extends to 3D on the square grids. Figure \ref{fig:fcn3d-net} shows an example of the network structure used in this paper. The network follows and simplifies the hourglass shape from \cite{Long}. Layer conv1, conv2 and conv3 downsample the input map by $1/2^3$ sequentially. Layer deconv4a and deconv4b upsample the incoming map by $2^3$ respectively. The ReLU activation is deployed after each layer. The output objectness map ($\mathbf{o}^a$) and bounding box map ($\mathbf{o}^b$) are collected from the deconv4a and deconv4b layers respectively.

Similar to DenseBox, the objectness region $\mathcal{V}$ is denoted as the center region of the object. For the proposed 3D case, a 3D sphere located at the object center is used. Points inside the sphere are labeled as positive / foreground label. The bounding box prediction at point $\mathbf{p}$ is encoded by the coordinate offsets, defined as:
\begin{equation}
    \Delta \mathbf{b}_\mathbf{p} = (\mathbf{c}_{\mathbf{p}, 1}^\top, \mathbf{c}_{\mathbf{p}, 2}^\top, \dots, \mathbf{c}_{\mathbf{p}, 8}^\top)^\top - (\mathbf{p}^\top, \dots, \mathbf{p}^\top)
\end{equation}
where $\mathbf{c}_{\mathbf{p}, \star}$ define the 3D coordinates of 8 corners of the object bounding box corresponding to the region $\mathbf{p}$.

The training and testing processes of the 3D CNN follows \cite{Li2016}. For the testing phase, candidate bounding boxes are extracted from regions predicted as objects and scored by counting its neighbors from all candidate bounding boxes. Bounding boxes are selected from the highest score and candidates overlapping with selected boxes are suppressed.

Figure \ref{fig:medium-output} shows an example of the detection intermediate results. Bounding box predictions from objectness points are plotted as green boxes. Note that for severely occluded vehicles, the bounding boxes shape are distorted and not clustered. This is mainly due to the lack of similar samples in the training phase.

\subsection{Comparison with 2D CNN}
Compared to 2D CNN, the dimension increment of 3D CNN inevitably consumes more computational resource, mainly due to 1) the memory cost of 3D data embedding grids and 2) the increasing computation cost of convolving 3D kernels.

On the other hand, naturally embedding objects in 3D space avoids perspective distortion and scale variation in the 2D case. This make it possible to learn detection using a relatively simpler network structure. 

\section{EXPERIMENTS}

We evaluate the proposed 3D CNN on the vehicle detection task from the KITTI benchmark \cite{Geiger2012}. The task contains images aligned with point cloud and object info labeled by both 3D and 2D bounding boxes.

The experiments mainly focus on detection of the \textit{Car} category for simplicity. Regions within the 3D center sphere of a \textit{Car} are labeled as positive samples, i.e. in $\mathcal{V}$. \textit{Van} and \textit{Truck} are labeled to be ignored. \textit{Pedestrian}, \textit{Bicycle} and the rest of the environment are labeled as negative background, i.e. $\mathcal{P} - \mathcal{V}$.

The KITTI training dataset contains 7500+ frames of data, of which 6000 frames are randomly selected for training in the experiments. The rest 1500 frames are used for offline validation, which evaluates the detection bounding box by its overlap with groundtruth on the image plane and the ground plane. The detection results are also compared on the KITTI online evaluation, where only the image space overlap are evaluated.

The KITTI benchmark divides object samples into three difficulty levels. Though this is is originally designed for the image based detection, we find that these difficulty levels can also be approximately used in difficulty division for detection and evaluation in 3D. The minimum height of 40px for the easy level approximately corresponds to objects within 28m and the minimum height of 25px for the moderate and hard levels approximately corresponds to object within 47m.

\subsection{Performance Analysis}

The original KITTI benchmark assumes that detections are presented as 2D bounding boxes on the image plane. Then the overlap area of the image plane bounding box with its ground truth is measured to evaluate the detection. However, from the perspective of building a complete autonomous driving system, evaluation in the 2D image space does not well reflect the demand of the consecutive modules including planning and control, which usually operates in world space, e.g. in the full 3D space or on the ground plane. Therefore, in the offline evaluation, we validate the proposed approach in both the image space and the world space, using the following metrics:

\begin{itemize}
    \item \textbf{Bounding box overlap on the image plane}. This is the original metric of the KITTI benchmark. The 3D bounding box detection is projected back to the image plane and the minimum rectangle hull of the projection is taken as the 2D bounding boxes. Some previous point cloud based detection methods \cite{Wang, Li2016, Behley2013a} also use this metric for evaluation. A detection is accepted if the overlap area IoU with the groundtruth is larger than 0.7.
    
    \item \textbf{Bounding box overlap on the ground plane}. The 3D bounding box detection is projected onto the 2D ground plane orthogonally. A detection is accepted if the overlap area IoU with the groundtruth is larger than 0.7. This metric reflects the demand of the autonomous driving system naturally, in which the vertical localization of the vehicle is less important than the horizontal.
    
\end{itemize}
For the above metrics, the naive  Average  Precision  (AP)  and  the  Average  Orientation Similarity (AOS) are both evaluated.

\begin{figure*}
    \centering
    \begin{tabular}{ccccc}
        \includegraphics[scale=0.35, trim=70 40 50 30, clip, angle=90]{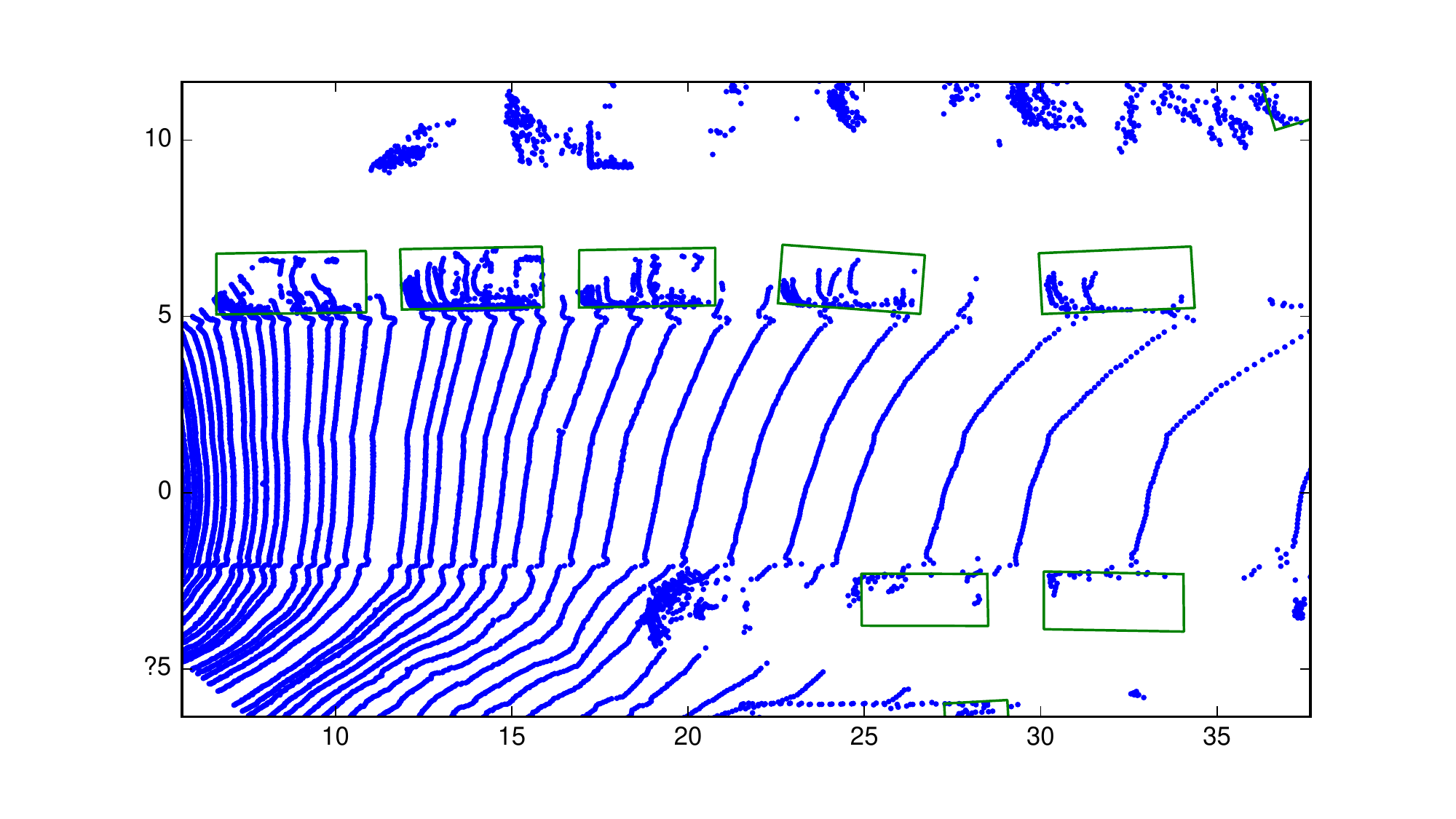} & \includegraphics[scale=0.35, trim=70 40 50 30, clip, angle=90]{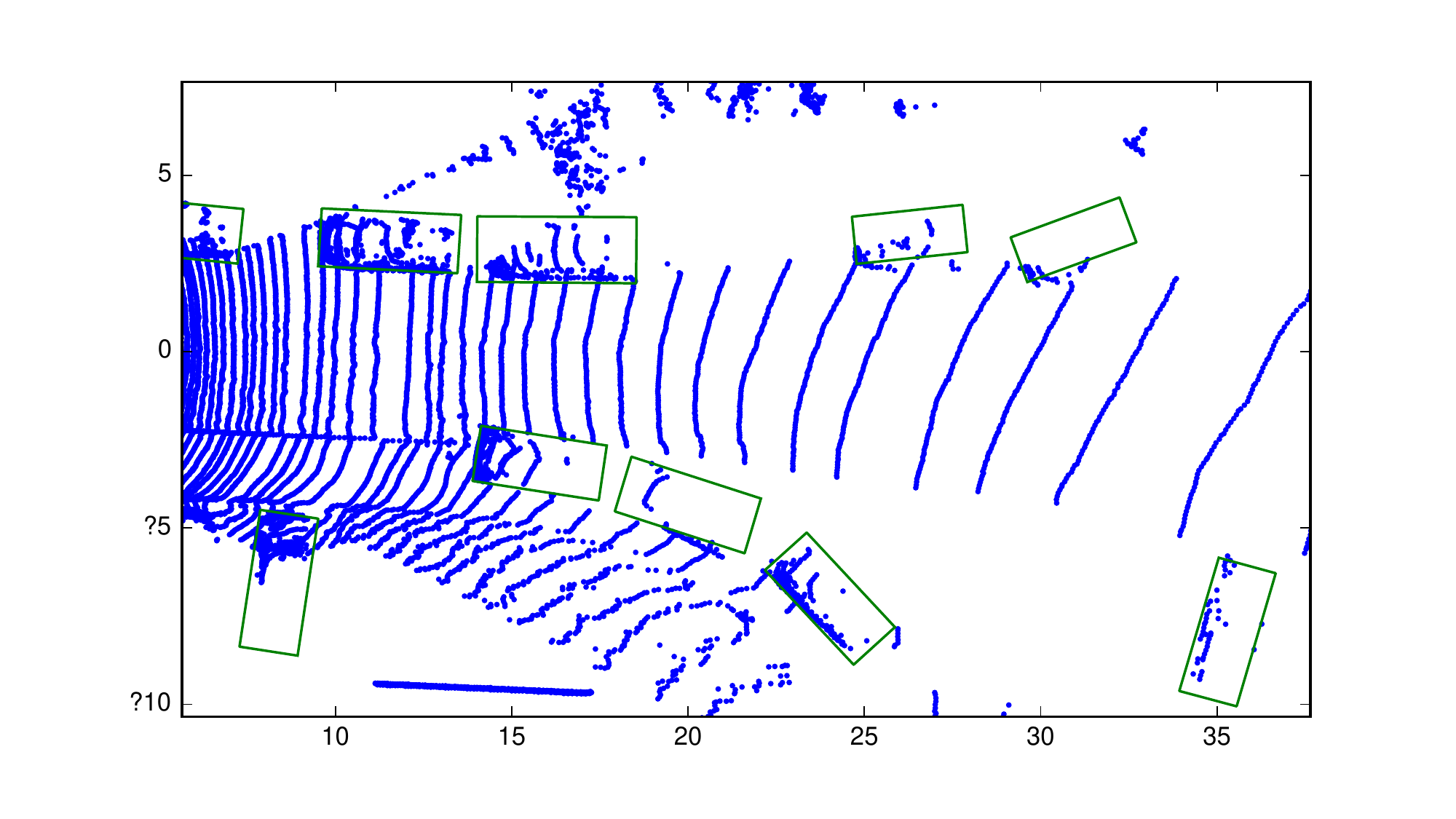} &
        \includegraphics[scale=0.35, trim=70 40 50 30, clip, angle=90]{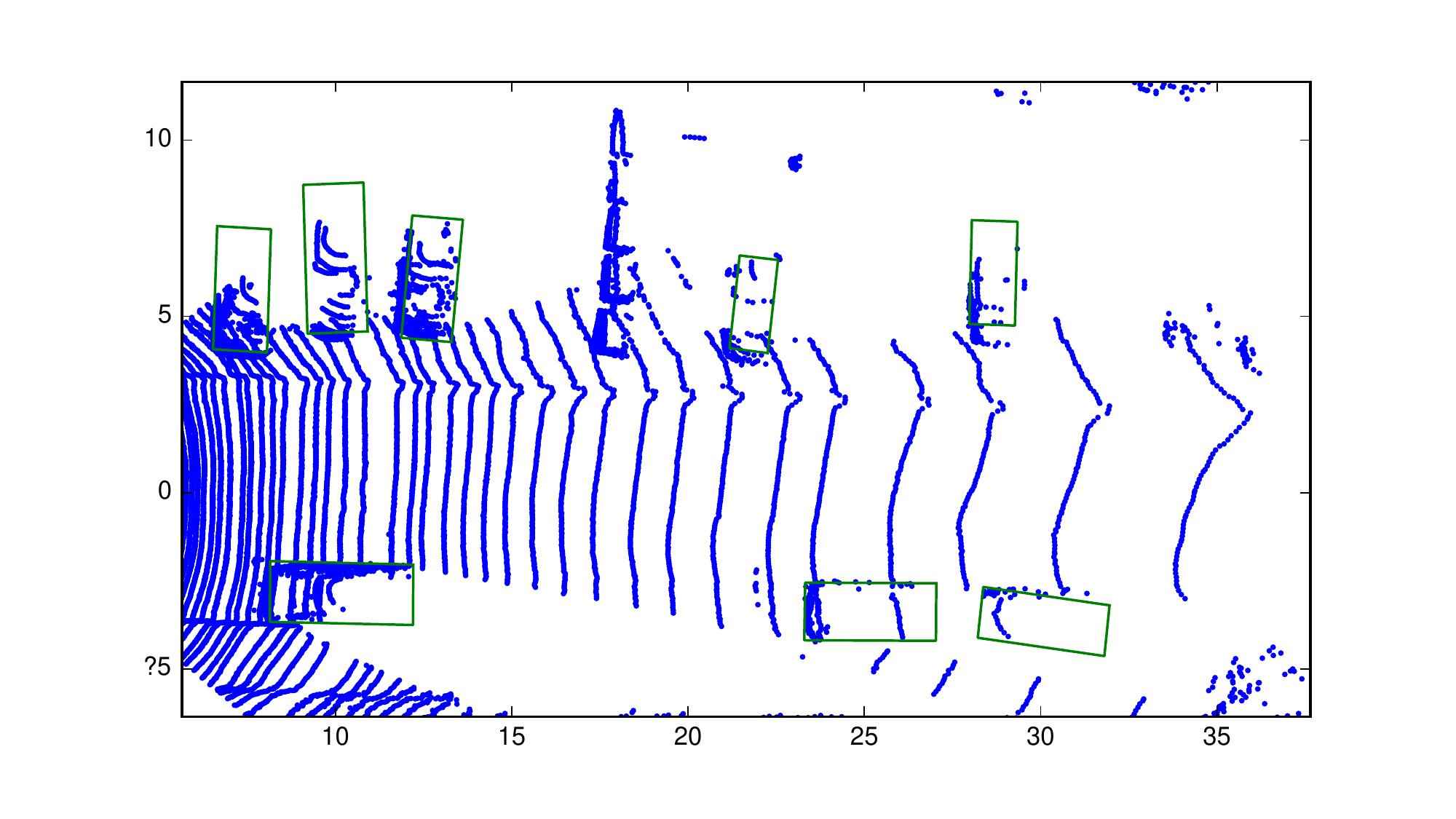} & \includegraphics[scale=0.35, trim=70 40 50 30, clip, angle=90]{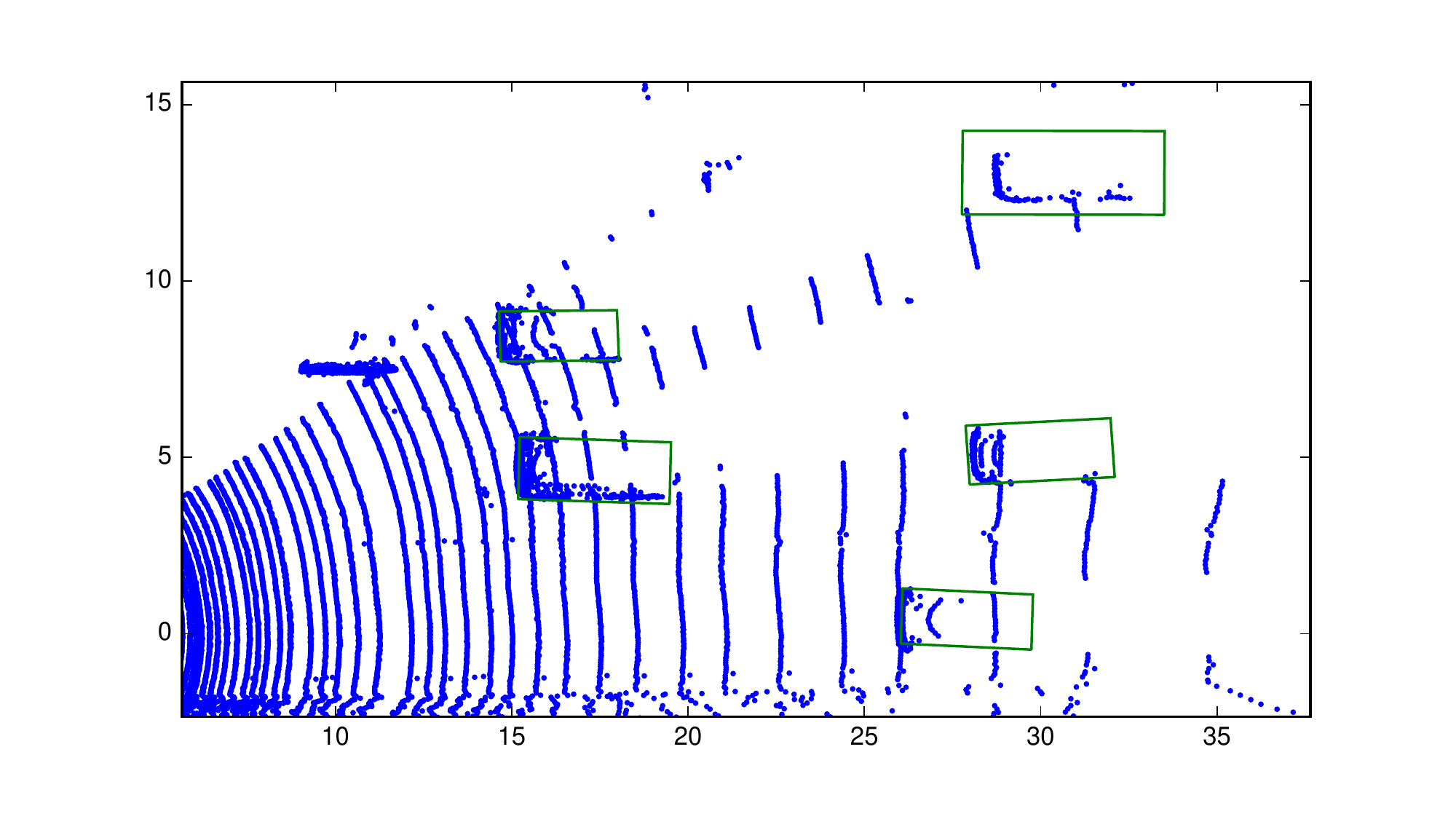} & \includegraphics[scale=0.35, trim=70 40 50 30, clip, angle=90]{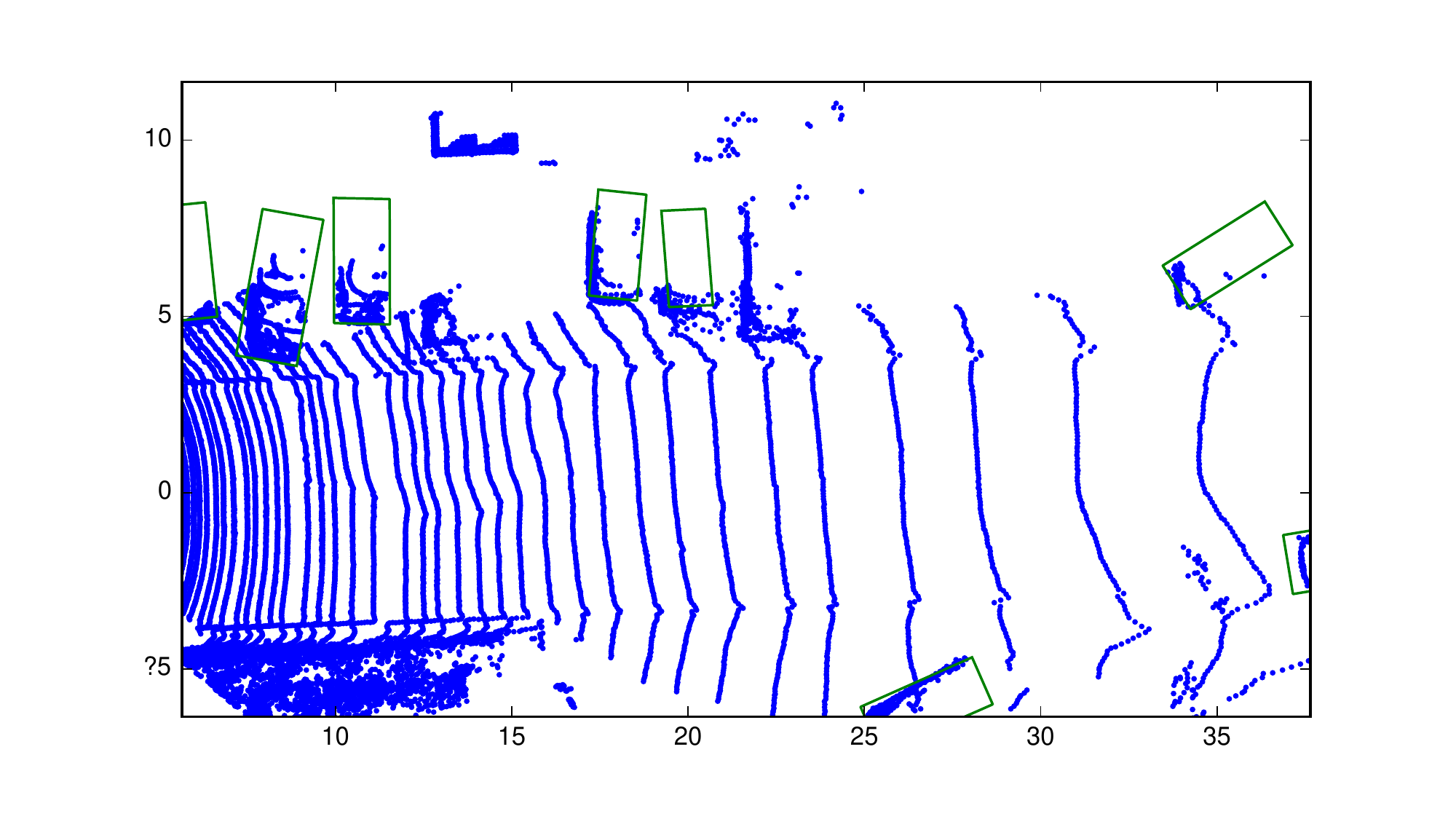}\\
        (a) & (b) & (c) & (d) & (e)
    \end{tabular}
    \caption{More detection results on the KITTI dataset using 3D FCN. }
    \label{fig:more-results}
\end{figure*}

The performance of the proposed approach and \cite{Li2016} is listed in Table \ref{tab:offline}. The proposed approach uses less layers and connections compared with \cite{Li2016} but achieves much better detection accuracy. This is mainly because objects have less scale variation and occlusion in 3D embedding. More detection results are visualized in Figure \ref{fig:more-results}.

\begin{table}
    \centering
    \caption{Performance in Average Precision and Average Orientation Similarity for the Offline Evaluation}
    \begin{tabular}{ccccc}
    \hline\hline
    & & Easy & Moderate & Hard \\
    \hline
    \multirow{2}{*}{Image Plane (AP)} & Proposed & 93.7\% & 81.9\% & 79.2\% \\
    & VeloFCN & 74.1\% & 71.0\% & 70.0\% \\
    \hline
    \multirow{2}{*}{Image Plane (AOS)} & Proposed & 93.7\% & 81.8\% & 79.1\%\\
    & VeloFCN & 73.9\% & 70.9\% & 69.9\%\\
    \hline
    \multirow{2}{*}{Ground Plane (AP)} & Proposed & 88.9\% & 77.3\% & 72.7\% \\
    & VeloFCN & 77.3\% & 72.4\% & 69.4\% \\
    \hline
    \multirow{2}{*}{Ground Plane (AOS)} & Proposed & 88.9\% & 77.3\% & 72.7\% \\
    & VeloFCN & 77.2\% & 72.3\% & 69.4\% \\
    \hline
    \end{tabular}
    \label{tab:offline}
\end{table}

\subsection{KITTI Online Evaluation}

The proposed approach is also evaluated on the KITTI online system. Note that on the current KITTI object detection benchmark image based detection algorithms outperforms previous point cloud based detection algorithms by a significant gap. This is due to two reasons: 1) The benchmark is using the metric of bounding box overlap on the image plane. Projecting 3D bounding boxes from point cloud inevitably introduce misalignment with 2D labeled bounding boxes. 2) Images have much higher resolution than point cloud (range scan), which enhances the detection of far or occluded objects. 

The proposed approach is compared with previous point cloud based detection algorithms and the results are listed in Table \ref{tab:online}. The performance of our method outperforms previous methods by a significant gap of $>20\%$, which is even comparable -- though not as well as yet -- with image based algorithms.

\begin{table}
    \centering
    \caption{Performance Comparison in Average Precision and Average Orientation Similarity for the KITTI Online Evaluation}
    \begin{tabular}{ccccc}
    \hline\hline
    & & Easy & Moderate & Hard \\
    \hline
    \multirow{4}{*}{Image Plane (AP)} & Proposed & 84.2\% & 75.3\% & 68.0\% \\
    & VeloFCN \cite{Li2016} & 60.3\% & 47.5\% & 42.7\% \\
    & Vote3D \cite{Wang} & 56.8\% & 48.0\% & 42.6\%\\
    & CSoR & 34.8\% & 26.1\% & 22.7\% \\
    & mBoW \cite{Behley2013a} & 36.0\% & 23.8\% & 18.4\% \\
    \hline
    \multirow{2}{*}{Image Plane (AOS)} & Proposed & 84.1\% & 75.2\% & 67.9\% \\
    & VeloFCN \cite{Li2016} & 59.1\% & 45.9\% & 41.1\% \\
    & CSoR & 34.0\% & 25.4\% & 22.0\% \\
    \hline
    \end{tabular}
    \label{tab:online}
\end{table}

\section{Conclusions}
Recent study in deploying deep learning techniques in point cloud have shown the promising ability of 3D CNN to interpret shape features. This paper attempts to further push this research. To the best of our knowledge, this paper proposes the first 3D FCN framework for end-to-end 3D object detection. The performance improvement of this method is significant compared to previous point cloud based detection approaches. While in this paper the framework are experimented on the point cloud collected by Velodyne 64E under the scenario of autonomous driving, it naturally applies to point cloud created by other sensors or reconstruction algorithms.

\section*{ACKNOWLEDGMENT}

The author would like to acknowledge the help from Xiaohui Li and Songze Li. Thanks also goes to Ji Wan and Tian Xia.

\bibliographystyle{plain}
\bibliography{root}

\end{document}